\documentclass{article}
\usepackage{spconf,amsthm,amsmath,amssymb,graphicx,hyperref}
\usepackage{multirow}
\usepackage{tabularx} 
\usepackage{booktabs}

\title{Null-LoRA: Low-Rank Adaptation on Null Space}
\name{Yi Zhang \qquad Yulei Kang \qquad Haoxuan Chen \qquad Jinxuan Li \qquad Jian-Fang Hu$^{*}$\thanks{* Corresponding author.}}
\address{School of Computer Science and Engineering, Sun Yat-sen University, Guangdong, China}
\begin{document}
%
\maketitle
\begin{abstract}
Parameter-efficient fine-tuning methods have gained considerable popularity for adapting large-scale models to downstream tasks, particularly LoRA and its variants. Existing methods perform low-rank adaptation over the full parameter space. However, fine-tuning within a subspace can achieve comparable effectiveness. Inspired by the observation that pre-trained models possess non-trivial null spaces, we propose \textbf{Null}-space based \textbf{Lo}w-\textbf{R}ank \textbf{A}daptation (Null-LoRA). Null-LoRA effectively reduces redundancy and enhances effective rank by freezing portions of the low-rank matrices. To further improve parameter efficiency, Null-LoRA constrains the entire incremental update within the null space, maximizing the utilization of incremental updates to adapt to new task paradigms. Null-LoRA surpasses the state of the art with fewer parameters in extensive experiments across image-text retrieval and visual question answering tasks.
\end{abstract}
\begin{keywords}
Parameter-Efficient Fine-Tuning, Vision-Language models, LoRA, Null Space.
\end{keywords}
\section{Introduction}
\label{sec:intro}

Large-scale pre-trained models have demonstrated remarkable capabilities in various fields including vision \cite{oquab2023dinov2}, text \cite{touvron2023llama}, and multi-modality \cite{li2022blip}. These models are first pre-trained with extensive data and then fully fine-tuned for each downstream task, adapting to different modalities and requirements to achieve superior performance in specific tasks or domains. This approach has been proven effective through numerous studies \cite{luo2022clip4clip, wang2023image}.


However, as the size of foundational models become increasingly large, large-scale models exhibit remarkable capabilities while containing billions or even hundreds of billions of parameters. The massive parameter count imposes substantial computational and storage costs, hindering the scalability of large models across diverse scenarios and tasks. To address these challenges, parameter pruning methods have been explored and proposed. Parameter-Efficient Fine-Tuning (PEFT) \cite{mangrulkar2022peft}, which freezes the backbone of large-scale models and introduces a small number of additional trainable parameters for fine-tuning, has gained prominence as a computationally efficient strategy. Among these methods, LoRA has seen widespread adoption due to its balanced performance and efficiency.
It is a prevalent phenomenon of pre-trained weights to exhibit varying degrees of rank deficiency, as clearly evidenced in models such as CLIP \cite{radford2021learning}, BLIP \cite{li2022blip}, and LLaMA \cite{touvron2023llama}. This non-full-rank property in pre-training introduces parameter redundancy and insufficient learning. Therefore, we propose that leveraging the intrinsic null space within pre-trained weights for updates helps models compensate for these inherent deficiencies. Concurrently, some studies \cite{yu2024language, panda2406lottery} have shown significant redundancy in LoRA's incremental updates. Maximizing the parameter efficiency of low-rank weights by reducing redundancy while enhancing update effectiveness becomes an important challenge.

\begin{figure*}[h]
    \centering
    \includegraphics[width=\textwidth]{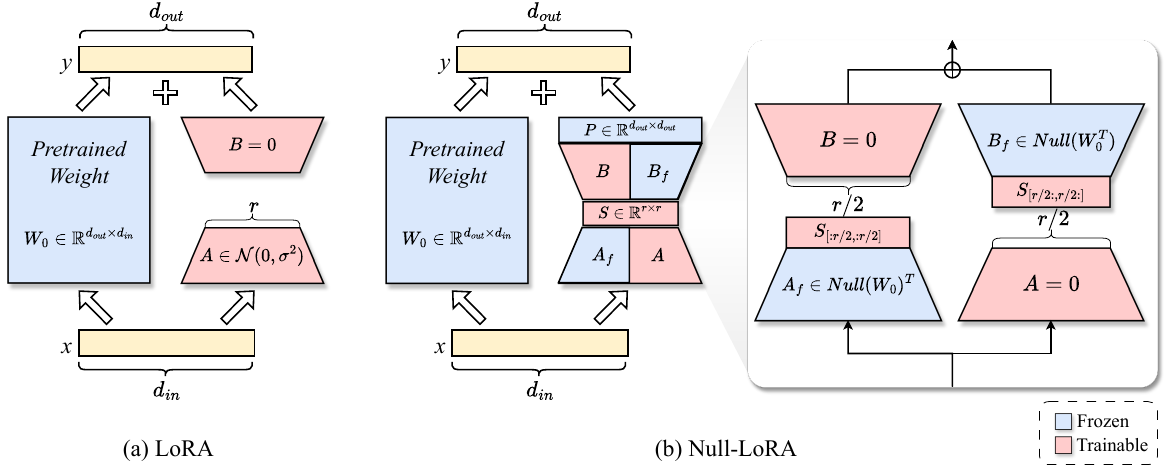}
    \caption{An overview of our proposed Null-LoRA. Note that $Null(\cdot)$ donates the right null space of a matrix.}
    \label{fig:Null-LoRA}
    \vspace{-3mm}
\end{figure*}

To address this challenge, we propose \textbf{Null}-space based \textbf{Lo}w-\textbf{R}ank \textbf{A}daptation (Null-LoRA). We split the low-rank matrices $B$ and $A$ into halves along the rank dimension, freezing one half of each. The frozen and trainable halves are then cross-paired and multiplied to form a low-rank decomposition product, as illustrated in Figure \ref{fig:Null-LoRA} (b). This ensures functional integrity while reducing parameter redundancy, thus enabling Null-LoRA to achieve higher intrinsic rank with reduced trainable parameters. To enhance update efficiency, we introduce a null-space constraint, obtaining the corresponding null-space orthonormal basis via Singular Value Decomposition (SVD) on the pre-trained weights to initialize the frozen low-rank decomposition matrices. Subsequently, through a null-space projection, Null-LoRA's entire incremental update is confined to the inactivate subspace within the pretrained weights. We evaluate Null-LoRA on multiple visual-language tasks, which obtains state-of-the-art performance compared to other PEFT methods. This demonstrates that Null-LoRA enhances parameter efficiency for low-rank matrices without increasing inference latency.

\section{Background and preliminary}
\label{sec:background}

While pre-trained models like LLaMA \cite{touvron2023llama}  hold broad linguistic knowledge, they are often not specialized for particular domains. To improve their performance on specific tasks while preserving their general language capabilities, users commonly fine-tune these models with task-relevant data.

Based on the hypothesis that updates during fine-tuning typically form a low "intrinsic rank", LoRA \cite{hu2022lora} fine-tunes a incremental update matrix as a product of two low-rank matrices to adapt models to downstream tasks with a significantly lower proportion of trainable parameters. Formally, given a pre-trained weight $W_0\in \mathbb{R}^{d_{out}\times d_{in}}$, the weight update is constrained to a low-rank decomposition:
\begin{equation}
y=W'x=W_0x+\Delta Wx=W_0x+BAx,
\end{equation}
where $W'\in \mathbb{R}^{d_{out}\times d_{in}}$ is the updated weight matrix, $B\in \mathbb{R}^{d_{out}\times r}$, $A\in \mathbb{R}^{r\times d_{in}}$ are the low-rank projection matrices, and $r\ll min\{d_{out}, d_{in}\}$ is the rank of projection matrices.

\section{The proposed method}
\label{sec:method}

\subsection{Cross-Freezing Projections with Norm Scaling}
\label{ssec:freeze}

\textbf{Cross-Freezing Low-Rank Projections.} Recent studies \cite{kopiczko2023vera, albert2025randlora} indicate that LoRA’s low-rank structure provides parameter efficiency but restricts representational capacity. Increasing the rank generally improves performance, yet also enlarges the number of trainable parameters, raising storage and computation costs. Meanwhile, other works \cite{yu2024language, panda2406lottery} have shown substantial redundancy in LoRA’s updates. This contradiction may partly stem from the fact that expanding rank while maintaining the same redundancy level increases the total number of parameters, allowing the parameter matrix to contain more effective updates. To address the above issues, inspired by \cite{zhang2023lora}, Null-LoRA freezes half of each low-rank matrix while training the other half, thereby increasing the effective rank and reducing redundancy without increasing computational costs. The weight update is defined as:
\begin{equation}
\Delta W =BA= 
\begin{pmatrix}
B \\
B_f
\end{pmatrix}
\begin{pmatrix}
A_f & A
\end{pmatrix}
=BA_f+B_fA,
\end{equation}
where $B\in \mathbb{R}^{d_{out}\times \frac{r}{2}}$, $A\in \mathbb{R}^{\frac{r}{2}\times d_{in}}$ are trainable halves and $B_f\in \mathbb{R}^{d_{out}\times \frac{r}{2}}$, $A_f\in \mathbb{R}^{\frac{r}{2}\times d_{in}}$ are frozen halves. Cross-freezing ensures that frozen down projections correspond to trainable up projections and vice versa, preserving structural effectiveness. Without this correspondence, the module reduces to LoRA of half rank with a fixed bias. Moreover, training only $B$ or $A$ may disrupt their complementary functions \cite{zhu2024asymmetry}, leading to incomplete functional adaptation. Hence, cross-freezing separate portions of $B$ and $A$ maintains the functional integrity of incremental updates.


\textbf{Dynamic Norm Scaling.} One can see that since $B_f$ and $A_f$ are orthogonal matrices, while $B$ and $A$ are unconstrained low-rank matrices, which leads to an imbalance in norms between them. The imbalanced weight norm will affect the gradient norm, which may in turn influence the model's generalization ability as studied in many works \cite{zhao2022penalizing}. To ensure that the two halves of the projection matrix have similar norms, Null-LoRA introduces a scaling vector $s\in \mathbb{R}^{r}$ to guide the alignment of matrix norms. The incremental update of the weight matrix is modified as:
\begin{equation}
\Delta W =BS_{[:\frac{r}{2}, :\frac{r}{2}]}A_f+B_fS_{[\frac{r}{2}:, \frac{r}{2}:]}A,
\end{equation}
where $S_{[:\frac{r}{2}, :\frac{r}{2}]}\in \mathbb{R}^{\frac{r}{2}\times \frac{r}{2}}$ and $S_{[\frac{r}{2}:, \frac{r}{2}:]}\in \mathbb{R}^{\frac{r}{2}\times \frac{r}{2}}$ are slices of the diagonal matrix formed by scaling vector $s$. By additionally learning the magnitude norms of the frozen matrix in each rank channel, Null-LoRA can improve computational stability, thereby enhancing the model's generalization ability.

\subsection{Null-Space Constraint}
\label{ssec:nullspace}

\textbf{Null-Space Initialization.} Existing pre-trained models widely exhibit non-full-rank phenomena, for instance, BLIP shows an average of 5\% rank deficiency in its linear layers. Null-LoRA initializes the frozen low-rank matrices $B_f$ and $A_f$ using null-space vectors from pre-trained weights, while setting the trainable matrices $B$ and $A$ to zero. Given a matrix $A \in \mathbb{R}^{m\times n}$, its null space is the set of vectors $x \in \mathbb{R}^n$ satisfying $Ax=0$. A basis matrix $B \in \mathbb{R}^{n\times r}$ constructed from linearly independent null-space vectors satisfies $AB=0$, where $r$ is the nullity (dimension of the null space). The rank and nullity of the matrices satisfy the identity:
\begin{equation}
rank(A)+rank(B)=rank(A)+nullity(A)=n.
\end{equation}
Consider $h$ as the output of multiplying the input $x$ by the updated weight $W'$. The square of its norm is formulated as:
\begin{equation}
\begin{aligned}
\Vert W'x \Vert^2
&=x^TW'^TW'x \\
&=x^T(W_0^T+\Delta W^T)(W_0+\Delta W)x.
\end{aligned}
\end{equation}
If $\Delta W$ can be expressed as a linear combination of a set of null-space basis vectors of $W^T$, then $\Delta W$ is in the null space of $W^T$, hence yielding $W_0^T\Delta W = 0$. Based on the above assumption, the squared norm of $h$ can be formulated as:
\begin{equation}
\begin{aligned}
\Vert W'x \Vert^2&=x^T(W_0^TW_0+\Delta W^T\Delta W)x \\
&=\Vert W_0x \Vert^2+\Vert \Delta Wx \Vert^2.
\end{aligned}
\end{equation}
That is, the squared norm of the updated output can be expressed as the sum of those of the pre-trained output and the incremental output, which implies that the two outputs are orthogonal. By constraining the fine-tuning update to be orthogonal to the pre-trained weight's direction, it is possible to explore a new orthogonal subspace while preserving that of the original pre-trained model, thus avoids forgetting the pre-trained latent knowledge during downstream task adaptation. 
To ensure that the assumption $W_0^T \Delta W = 0$ holds, the following constraint is introduced:
\begin{equation}
\begin{aligned}
W_0^T\Delta W
&=W_0^TBS_{[:\frac{r}{2}, :\frac{r}{2}]}A_f+W_0^TB_fS_{[\frac{r}{2}:, \frac{r}{2}:]}A=0.
\end{aligned}
\end{equation}
Following existing methods on null-space projection \cite{wang2021training}, Null-LoRA applies Singular Value Decomposition (SVD) to the pre-trained weight $W_0$:
\begin{equation}
U,\Sigma,V^T=\text{SVD}(W_0).
\end{equation}
Let $\hat{U}$ and $\hat{V}$ denote the submatrices of $U$ and $V$ corresponding to zero singular values, with $\hat{U}\in \mathbb{R}^{d_{out}\times r}$ and $\hat{V}\in \mathbb{R}^{r\times d_{in}}$. These submatrices form bases of the left and right null spaces of $W_0$, satisfying $W_0^T \hat{U}=0$ and $W_0 \hat{V}=0$. Null-LoRA uses $\hat{U}$ and $\hat{V}$ to initialize the frozen low-rank matrices $B_f$ and $A_f$. Notably, such initialization ensures $W_0^T B_f S_{[\frac{r}{2}:,\frac{r}{2}:]} A = 0$, but does not guarantee $W_0^T B S_{[:\frac{r}{2},:\frac{r}{2}]} A_f = 0$ since $B$ is unconstrained. To address this, Null-LoRA introduces a projection matrix $P$, defined as:
\begin{equation}
P=\hat{U}\hat{U}^T.
\end{equation}
Projection matrix $P$ projects the incremental update matrix into the null space of $W_0^T$. Since $\hat{U}$ is a submatrix of $U$, $\hat{U}$ is also an orthogonal matrix. Furthermore, by the properties of orthogonal matrices, $\hat{U}$ satisfies $\hat{U}^T \hat{U} = I$. Thus we have $PB_f = \hat{U}\hat{U}^T\hat{U} = \hat{U}I = B_f$. The final incremental update of Null-LoRA is then formed as:
\begin{equation}
\begin{aligned}
\Delta W
&=PBS_{[:\frac{r}{2},:\frac{r}{2}]}A_f+PB_fS_{[\frac{r}{2}:,:\frac{r}{2}:]}A \\
&=PBS_{[:\frac{r}{2},:\frac{r}{2}]}A_f+B_fS_{[\frac{r}{2}:,:\frac{r}{2}:]}A.
\end{aligned}
\end{equation}
By projecting the column vectors of $B$ into the null space of $W_0^T$ using the projection matrix $P$, the incremental update is restricted to lie within the null space of $W_0^T$. The constraint $W_0^T\Delta W=0$ ensures that Null-LoRA's fine-tuning update direction remains orthogonal to the pre-training direction, thereby decoupling pre-training from fine-tuning update and achieving optimal utilization of incremental update to adapt to new task paradigms.

\textbf{Rank Self-Adaptation.} Since Null-LoRA utilizes the null space of pretrained weights to initialize low-rank projection matrices, the rank of each layer is adaptively determined by the rank deficiency of its corresponding pretrained weight. Specifically, the rank of each incremental update is set to twice the nullity of its corresponding pre-trained weight, thereby allocated to the two halves of the low-rank matrix. Through self-adaptation, Null-LoRA assigns higher ranks to layers with greater rank deficiencies to enhance their performance, while assigning lower ranks to layers with lesser deficiencies to reduce overall computational cost. Constructing incremental update using the inactive subspaces within non-full-rank weights also improves model's representation power for downstream tasks. 

\begin{table*}[t!]
  \centering
  \caption{Results on image-text retrieval datasets MSCOCO and FLICKR30K. Using text queries to retrieve images is simplified as T→I and vice versa. \# Tunable is the number of tunable parameters. The best result among the methods with frozen backbone is marked in bold, and the second best in underline. All performances of compared methods are obtained from \cite{guo2025parameter}.}
  \footnotesize  
  \resizebox{\textwidth}{!}{
  \begin{tabular}{@{\hspace{1.5mm}\extracolsep{\fill}}lr|cccccc|cccccc@{\hspace{1.5mm}}}
    \toprule
    \multirow{2}{*}{Method} & \multirow{2}{*}{\# Tunable} & 
    \multicolumn{3}{c}{MSCOCO (I→T)} & \multicolumn{3}{c|}{MSCOCO (T→I)} & 
    \multicolumn{3}{c}{FLICKR30K (I→T)} & \multicolumn{3}{c}{FLICKR30K (T→I)} \\
    & & R@1 & R@5 & R@10 & R@1 & R@5 & R@10 & R@1 & R@5 & R@10 & R@1 & R@5 & R@10 \\
    \midrule
    \multicolumn{14}{@{\hspace{1.5mm}\extracolsep{\fill}}l}{\textbf{Methods with full fine-tuning}} \\
    \cmidrule{1-14}
    UNITER & 330M & 65.7 & 88.6 & 93.8 & 52.9 & 79.9 & 88.0 
    & 87.3 & 98.0 & 99.2 & 75.6 & 94.1 & 96.8 \\
    VILLA & 330M & - & - & - & - & - & - 
    & 87.9 & 97.5 & 98.8 & 76.3 & 94.2 & 96.8 \\
    OSCAR & 330M & 73.5 & 92.2 & 96.0 & 57.5 & 82.8 & 89.8 
    & - & - & - & - & - & - \\
    ALIGN & 820M & 77.0 & 93.5 & 96.9 & 59.9 & 83.3 & 89.8 
    & 95.3 & 99.8 & 100.0 & 84.9 & 97.4 & 98.6 \\
    ALBEF & 210M & 77.6 & 94.3 & 97.2 & 60.7 & 84.3 & 90.5 
    & 95.9 & 99.8 & 100.0 & 85.6 & 97.5 & 98.9 \\
    BLIP & 223M & 81.9 & 95.4 & 97.8 & 64.3 & 85.7 & 91.5
    & 97.3 & 99.9 & 100.0 & 87.3 & 97.6 & 98.9 \\

    \cmidrule{1-14}
    \multicolumn{14}{@{\hspace{1.5mm}\extracolsep{\fill}}l}{\textbf{Methods with frozen backbone}} \\
    \cmidrule{1-14}
    
    LoRA (r=32) & 10.6M & 76.7 & 91.2 & 96.0 & 60.1 & 82.2 & 89.5 
    & 96.3 & 99.7 & \underline{99.8} & 84.8 & 96.6 & 98.4 \\
    DoRA (r=32) & 10.8M & 77.8 & 93.5 & 96.6 & 61.1 & 83.9 & 90.1 
    & \textbf{96.7} & \underline{99.8} & \textbf{100.0} & 85.1 & \underline{97.2} & 98.5 \\
    UniAdapter (r=512) & 19.5M & \underline{79.6} & \underline{94.5} & \textbf{97.3} & \underline{62.5} & \textbf{85.0} & \underline{91.0} 
    & \textbf{96.7} & 99.7 & \textbf{100.0} & \underline{86.2} & \textbf{97.3} & \textbf{98.8} \\
    Aurora (r=64) & 0.6M & 78.0 & 93.4 & 96.7 & 61.5 & 84.0 & 90.4 
    & \textbf{96.7} & \underline{99.8} & \textbf{100.0} & 85.8 & \underline{97.2} & \underline{98.7} \\
    Null-LoRA (ours) & 6.0M & \textbf{80.7} & \textbf{94.6} & \underline{97.2} & \textbf{62.7} & \underline{84.9} & \textbf{91.1} 
    & \underline{96.6} & \textbf{100.0} & \textbf{100.0} & \textbf{86.3} & \textbf{97.3} & \textbf{98.8} \\
    \bottomrule
  \end{tabular}
  }


  \label{table:retrieval}
  \vspace{-4mm}
\end{table*}

\section{Experiments}
\label{sec:experiments}

\subsection{Experimental Settings}
\label{ssec:setting}

\textbf{Datasets \& Baselines.} We evaluate our Null-LoRA on 3 cross-modal downstream datasets, including image-text retrieval datasets: MSCOCO \cite{lin2014microsoft} and Flickr30K \cite{plummer2015flickr30k}; and visual question answering (VQA) dataset: VQAv2 \cite{goyal2017making}, collectively covering two distinct cross-modal tasks. We compare Null-LoRA with various baselines, including full fine-tuning approaches and multiple frozen backbone methods including LoRA \cite{hu2022lora}, DoRA \cite{liu2024dora}, Aurora \cite{wang2023parameter} and UniAdapter \cite{lu2023uniadapter}.

\textbf{Implementation Details.} We apply BLIP-base \cite{li2022blip} as our visual-language backbone for all downstream tasks, while keeping the parameters of the backbone model frozen during the fine-tuning process. We use AdamW \cite{loshchilov2017decoupled} optimizer with a weight decay of 0.05 and a learning rate of 1e-4 for all experiments. All experiments are conducted on 8$\times$ NVIDIA A6000 GPU (48G).

\begin{table}[t]
  \centering
  \caption{Results on VQA dataset VQAv2.}
  \small
  \begin{tabular}{@{\hspace{1.5mm}\extracolsep{\fill}}lr|cc@{\hspace{1.5mm}}}
    \toprule
    \multirow{2}{*}{Method} & \multirow{2}{*}{\# Tunable} & 
    \multicolumn{2}{c}{VQAv2} \\
    & & test-dev & test-std \\
    \midrule
    \multicolumn{4}{@{\hspace{1.5mm}\extracolsep{\fill}}l}{\textbf{Methods with full fine-tuning}} \\
    \cmidrule{1-4}
    VL-T5/BART & 165M & - & 71.30 \\
    SOHO & 155M & 73.25 & 73.47 \\
    OSCAR & 330M & 73.61 & 73.82 \\
    UNITER & 330M & 73.82 & 74.03 \\
    ALBEF & 266M & 75.84 & 76.04 \\
    BLIP & 337M & 77.44 & 77.48 \\
    \cmidrule{1-4}
    \multicolumn{4}{@{\hspace{1.5mm}\extracolsep{\fill}}l}{\textbf{Methods with frozen backbone}} \\
    \cmidrule{1-4}
    LoRA (r=32) & 10.6M & 75.10 & 75.20 \\
    DoRA (r=32) & 10.8M & 75.89 & 76.17 \\
    UniAdapter (r=512) & 19.5M & 75.44 & 75.56 \\
    Null-LoRA (ours) & 9.5M & \textbf{77.48} & \textbf{77.48} \\
    \bottomrule
  \end{tabular}
  \label{table:vqa}
  \vspace{-4mm}
\end{table}

\subsection{Comparisons with State-Of-The-Art}
\label{ssec:comparison}

\textbf{Image-Text Retrieval.} Table \ref{table:retrieval} shows performances on image-text retrieval tasks on MSCOCO \cite{lin2014microsoft} and FLICKR30K \cite{plummer2015flickr30k}. It can be observed that our Null-LoRA achieves better results than LoRA and DoRA while using fewer parameters. Compared to parameter sharing methods, Null-LoRA also achieves comparable performance using 30\% of the trainable parameters of UniAdapter. Although Aurora has a smaller parameter count, its performance lags behind other methods.

\textbf{Visual Question Answering.} We further evaluate performances on VQAv2 \cite{goyal2017making} for visual question answering task and demonstrate the results in Table \ref{table:vqa}. Unlike image-text retrieval tasks, pre-trained weights used for VQA task exhibit greater rank deficiencies, introducing more trainable parameters. Null-LoRA achieves superior performance compared to other methods, indicating our approach's strong comprehension capabilities in visual-language reasoning scenarios.

\subsection{Analysis of Different Designs}
\label{ssec:analysis}

\textbf{Effectiveness of Cross-Freezing.} By examining the intrinsic rank of fine-tuned weights, it is observed that the stacked projection matrix are nearly full-rank. This implies that by fixing half the rank of the projection matrices, the remaining half adaptively learns linearly independent vectors from the frozen basis vector set. This demonstrates that Null-LoRA improves parameter efficiency by pre-determining a portion of the required rank through cross-freezing.

\textbf{Effectiveness of Null-Space.} Effectiveness of null-space projection is demonstrated in Table \ref{table:ablation}, where w/o null space indicates randomly initialized frozen matrices. By constraining incremental updates within the null-space, the model focuses on exploring the inactivate subspace, thereby achieving performance gains. Cross-freezing expands the effective rank, whereas null-space projection introduces intrinsic rank into effective subspace. These mechanisms interact synergistically to deliver better performance.


\begin{table}[tp]
  \centering
  \caption{Comparison results obtained with and without null space on the VQAv2 dataset.}
  \small
  \begin{tabular}{@{\hspace{1.5mm}\extracolsep{\fill}}lr|ccc@{\hspace{1.5mm}}}
    \toprule
    \multirow{2}{*}{Method} & \multirow{2}{*}{\# Tunable} & 
    \multicolumn{2}{c}{VQAv2} \\
    & & test-dev & test-std \\
    \midrule
    Null-LoRA & 9.5M & \textbf{77.48} & \textbf{77.48} \\
    w/o null space & 9.5M & 77.02 & 77.04 \\
    \bottomrule
  \end{tabular}
  \label{table:ablation}
  \vspace{-4mm}
\end{table}

\section{Conclusion}
\label{sec:conclusion}

We introduce Null-LoRA, a parameter-efficient fine-tuning method based on null space and cross-freezing. Null-LoRA employs partial freezing of low-rank matrices, cross-pairs them, and incorporates null-space constraints to reduce redundancy and enhance the effective rank, thereby improving parameter efficiency. Through extensive experiments Null-LoRA demonstrates effectiveness, achieving superior performance with fewer trainable parameters compared to current state-of-the-art methods.




\vfill\pagebreak




\bibliographystyle{IEEEbib}
\small
\bibliography{refs}

\end{document}